\relax
\documentclass[letterpaper]{article} % DO NOT CHANGE THIS
\usepackage[nonatbib,preprint]{neurips_2020}

\usepackage{times}  % DO NOT CHANGE THIS
\usepackage{helvet} % DO NOT CHANGE THIS
\usepackage{courier}  % DO NOT CHANGE THIS
\usepackage[hyphens]{url}  % DO NOT CHANGE THIS
\usepackage{graphicx} % DO NOT CHANGE THIS
\usepackage{graphicx}  % DO NOT CHANGE THIS
\usepackage{algorithm}
\usepackage[noend]{algpseudocode}
\usepackage{booktabs}
\usepackage{import}
\usepackage{amsmath}
\usepackage[toc,page,header]{appendix}
\usepackage{minitoc}

\usepackage{wrapfig}
\usepackage[round]{natbib}
\usepackage[colorlinks=true,linkcolor=black,citecolor=blue]{hyperref}       % hyperlinks

\usepackage{amsmath}
\usepackage{mathtools}
\usepackage[bb=boondox,bbscaled=.95]{mathalfa}
\usepackage{tikz}
\usepackage{pgfplots}
\usepgfplotslibrary{groupplots}
\usepgfplotslibrary{patchplots}
\pgfplotsset{compat=1.14}

\usepackage{bm}

\DeclareMathOperator{\dom}{dom}

\DeclareMathOperator{\R}{\mathbb{R}}
\DeclareMathOperator{\Nat}{\mathbb{N}}

\newcommand{\parM}{\bm\Theta}
\newcommand{\norm}[1]{\left\lVert#1\right\rVert}

\DeclareMathOperator{\V}{\mathcal{V}}

\DeclareMathOperator{\C}{\mathcal{C}}

\DeclareMathOperator{\N}{\mathcal{N}}
\DeclareMathOperator{\K}{\mathcal{K}}
\DeclareMathOperator{\G}{\mathcal{G}}
\DeclareMathOperator{\E}{\mathcal{E}}
\DeclareMathOperator{\Sa}{\mathcal{S}}
\DeclareMathOperator{\T}{\mathcal{T}}
\DeclareMathOperator{\I}{\mathcal{I}}

\newcommand{\h}{\mathbf{h}}
\newcommand{\f}{\mathbf{f}}
\newcommand{\x}{\mathbf{x}}

\newcommand{\yb}{\mathbf{y}}
\newcommand{\y}{\mathbf{y}}
\newcommand{\g}{\mathbf{g}}
\newcommand{\ub}{\mathbf{u}}
\newcommand{\m}{\mathbf{m}}

\newcommand{\Ab}{\mathbf{A}}

\newcommand{\Fb}{\mathbf{F}}
\newcommand{\Hb}{\mathbf{H}}

\newcommand{\Lb}{\mathbf{L}}

\newcommand{\Xb}{\mathbf{X}}
\newcommand{\Yb}{\mathbf{Y}}

\newcommand{\lc}{\left(}
\newcommand{\rc}{\right)}

\tikzset{
  NNnode/.pic={
  \pgfmathsetmacro\RecH{2}
  \pgfmathsetmacro\RecW{\RecH/10}
  \coordinate (-ll) at (-\RecW/2,-\RecH/2);
  \coordinate (-ur) at (\RecW/2,\RecH/2);
  \coordinate (-lr) at (-ll-|-ur);
  \coordinate (-ul) at (-ll|--ur);
  \path (-ul) -- (-ur) coordinate[midway] (-north);
  \path (-ll) -- (-lr) coordinate[midway] (-south);
  \path (-ll) -- (-ul) coordinate[midway] (-west);
  \path (-ur) -- (-lr) coordinate[midway] (-east);

  \begin{scope}[shift={(-\RecW/2,-\RecH/2)}]
  \draw (-ll) rectangle (-ur);
  \foreach \y in {0.05,0.5,0.75,0.85,0.95}
    \draw (0.5*\RecW,\RecH*\y) circle[radius=0.3*\RecW];
  \foreach \y in {0.275,0.625} {
    \fill (\RecW*0.4,\y*\RecH-0.1*\RecW) rectangle (0.6*\RecW,\y*\RecH-0.3*\RecW);
    \fill (\RecW*0.4,\y*\RecH+0.1*\RecW) rectangle (0.6*\RecW,\y*\RecH+0.3*\RecW);
    }
  \end{scope}
  }
}

\usepackage[most]{tcolorbox}

\definecolor{olive}{rgb}{0.6, 0.6, 0.2}
\definecolor{sand}{rgb}{0.8666666666666667, 0.8, 0.4666666666666667}
\definecolor{wine}{rgb}{0.5333333333333333, 0.13333333333333333, 0.3333333333333333}
\definecolor{deblue}{RGB}{11,132,147}
\definecolor{ocra}{RGB}{204, 119, 34}
\definecolor{depurple}{RGB}{131, 102, 135}
\definecolor{degrey}{RGB}{186, 172, 172}
\newtcolorbox{CatchyBox}[2][]{
    lower separated=false,
    colback=white!80!degrey!90!depurple,
    colframe=white, fonttitle=\bfseries,
    colbacktitle=white!50!degrey!90!depurple,
    coltitle=black,
    enhanced,
    attach boxed title to top left={xshift=.02\linewidth,yshift=-4mm},
    title=#2,#1}

\title{Graph Neural Ordinary Differential Equations}
\author{Michael Poli\textsuperscript{* \rm 1},  Stefano Massaroli\textsuperscript{* \rm 2}, Junyoung Park\textsuperscript{\rm 1}\\ \textbf{Atsushi Yamashita\textsuperscript{\rm 2}, Hajime Asama\textsuperscript{\rm 2}, Jinkyoo Park\textsuperscript{\rm 1}} \\ % All authors must be in the same font size and format. Use \Large and \mathbf to achieve this result when breaking a line
\textsuperscript{\rm 1}Department of Industrial \& Systems Engineering, KAIST, Daejeon, South Korea \\
\textsuperscript{\rm 2}Department of Precision Engineering, The University of Tokyo, Tokyo, Japan \\
{\tt \{poli\_m, junyoung, jinkyoo.park\}@kaist.ac.kr,}\\{ \tt \{massaroli, yamashita, asama\}@robot.t.u-tokyo.ac.jp}% email address must be in roman text type, not monospace or sans serif 
}
\begin{document}
\maketitle

\begin{abstract}
We introduce the framework of continuous--depth \textit{graph neural networks} (GNNs). \textit{Graph neural ordinary differential equations} (GDEs) are formalized as the counterpart to GNNs where the input--output relationship is determined by a \textit{continuum} of GNN layers, blending discrete topological structures and differential equations. The proposed framework is shown to be compatible with various static and autoregressive GNN models. Results prove general effectiveness of GDEs: in static settings they offer computational advantages by incorporating numerical methods in their forward pass; in dynamic settings, on the other hand, they are shown to improve performance by exploiting the geometry of the underlying dynamics.
\end{abstract}
\doparttoc
\faketableofcontents
\section{Introduction}
Introducing appropriate inductive biases on deep learning models is a well--known approach to improving sample efficiency and generalization performance \citep{battaglia2018relational}. Graph neural networks (GNNs) represent a general computational framework for imposing such inductive biases when the problem structure can be encoded as a graph or in settings where prior knowledge about entities composing a target system can itself be described as a graph \citep{li2018combinatorial,gasse2019exact,sanchez2018graph}. 

\begin{wrapfigure}[18]{r}{0.5\textwidth}
    \centering
    \vspace{-6mm}
    \includegraphics[scale=0.75]{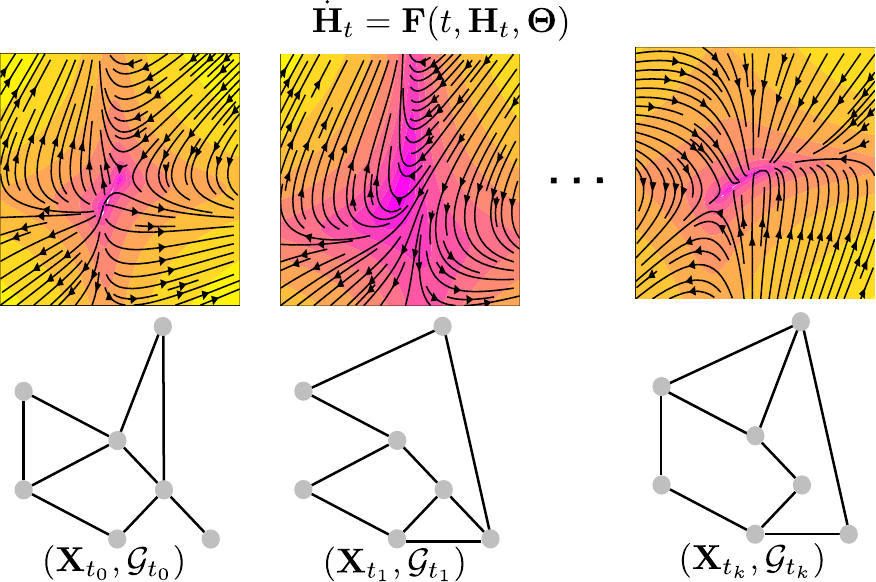}
    \caption{\textit{Graph neural ordinary differential equations} (GDEs) model vector fields defined on graphs, both in cases when the structure is fixed or changes in time, via a continuum of \textit{graph neural network} (GNN) layers.}
    \label{fig:vek}
\end{wrapfigure}GNNs have shown remarkable results in various application areas such as node classification \citep{zhuang2018dual,gallicchio2019fast}, graph classification \citep{yan2018spatial} and forecasting \citep{li2017diffusion,wu2019graph} as well as generative tasks \citep{li2018learning,you2018graphrnn}.
A different but equally important class of inductive biases is concerned with the type of temporal behavior of the systems from which the data is collected i.e., discrete or continuous dynamics. Although deep learning has traditionally been a field dominated by discrete models, recent advances propose a treatment of neural networks equipped with a continuum of layers \citep{haber2017stable,chen2018neural}. This view allows a reformulation of the forward and backward pass as the solution of the initial value problem of an \textit{ordinary differential equation} (ODE). Such approaches allow direct modeling of ODEs and can guide  discovery of novel general purpose deep learning models.
\newpage
\paragraph{Blending graphs and differential equations}
In this work we propose the system--theoretic framework of \textit{graph neural ordinary differential equations} (GDEs) by defining ODEs parametrized by GNNs. GDEs are designed to inherit the ability to impose relational inductive biases of GNNs while retaining the dynamical system perspective of continuous--depth models. We validate GDEs experimentally on a static semi--supervised node classification task as well as spatio--temporal forecasting tasks. GDEs are shown to outperform their discrete GNN analogues: the different sources of performance improvement are identified and analyzed separately for static and dynamic settings.

\paragraph{Sequences of graphs}
We extend the GDE framework to the spatio--temporal setting and formalize a general autoregressive GDE model as a \textit{hybrid dynamical system}. The structure--dependent vector field learned by GDEs offers a data--driven approach to the modeling of dynamical networked systems \citep{lu2005time,andreasson2014distributed}, particularly when the governing equations are highly nonlinear and therefore challenging to approach with analytical methods. Autoregressive GDEs can adapt the prediction horizon by adjusting the integration interval of the ODE, allowing the model to track the evolution of the underlying system from irregular observations.

\paragraph{GDEs as general--purpose models}
In general, no assumptions on the continuous nature of the data generating process are necessary in order for GDEs to be effective. Indeed, following recent work connecting different discretization schemes of ODEs \citep{lu2017beyond} to previously known architectures such as FractalNets \citep{larsson2016fractalnet}, we show that GDEs can equivalently be utilized as high--performance general purpose models. In this setting, GDEs offer a grounded approach to the embedding of classic numerical schemes inside the forward pass of GNNs.
\section{Graph Neural Ordinary Differential Equations}
We begin by introducing the general formulation of GDEs.
\subsection{General Framework}
\paragraph{Definition of GDE}
Without any loss of generality, the inter--layer dynamics of a GNN node feature matrix can be represented in the form:
\begin{equation*}
    \left\{
        \begin{matrix*}[l]
            \Hb{(s+1)} = \Hb(s) + \Fb_{\G}\lc s, \Hb(s), \parM(s)\rc \\
            \Hb(0) = \Xb_e
        \end{matrix*}
    \right.,~~s\in\Nat,
\end{equation*}
where $\Xb_e\in\R^{n\times h}$ is an embedding of $\Xb$\footnote{$\Xb_e$ can be obtained from $\Xb$, e.g. with a single linear layer: $\Xb_e := \Xb\mathbf{W}$, $\mathbf{W}\in\R^{d\times h}$ or with another GNN layer.}, $\Fb_{\G}$ is a matrix--valued nonlinear function conditioned on graph $\mathcal{G}$ and $\parM(s)\in\R^p$ is the tensor of trainable parameters of the  $s$-th layer. Note that the explicit dependence on $s$ of the dynamics is justified in some graph architectures, such as diffusion graph convolutions \citep{atwood2016diffusion}. 
A \textit{graph neural differential ordinary equation} (GDE) is defined as the following Cauchy problem:
\begin{equation}\label{eq:GDE}
    \left\{
        \begin{matrix*}[l]
            \dot\Hb(s) = \Fb_{\G}\lc s, \Hb(s), \parM\rc \\
            \Hb(0) = \Xb_e
        \end{matrix*}
    \right.,~~s\in\Sa\subset\R,
\end{equation}
where $\Fb_{\G}:\Sa\times\R^{n\times h}\times\R^p\rightarrow\R^{n\times h}$ is a depth--varying vector field defined on graph $\mathcal{G}$. 

To reduce stiffness of learned vector fields, alleviating the computational burden of adaptive ODE solvers, the node features can be augmented in several ways  \citep{dupont2019augmented,massaroli2020dissecting} by concatenating additional dimensions or prepending input layers to the GDE.
\paragraph{Well--posedness} Let $\Sa:=[0,1]$. Under mild conditions on $\Fb$, namely Lipsichitz continuity with respect to $\Hb$ and uniform continuity with respect to $s$,  for each initial condition (GDE embedded input) $\Xb_e$, the ODE in \eqref{eq:GDE} admits a unique solution $\Hb(s)$ defined in the whole $\Sa$. Thus there is a mapping $\bm\Psi$ from $\R^{n\times h}$ to the space of absolutely continuous functions $\Sa\to\R^{n\times h}$ such that $\Hb := \bm\Psi(\Xb_e)$ satisfies the ODE in \eqref{eq:GDE}. This implies the the output $\Yb$ of the GDE satisfies
\[
    \Yb = \bm\Psi(\Xb_e)(1).
\]
Symbolically, the output of the GDE is obtained by the following
\begin{equation*}
    \Yb = \Xb_e + \int_{\Sa}\Fb_{\G}(\tau,\Hb(\tau),\parM)d\tau.
\end{equation*}
Note that applying an output layer or network to $\Yb$ before passing it to downstream applications is generally beneficial.

\paragraph{Integration domain} We restrict the integration interval to $\Sa=[0,1]$, given that any other integration time can be considered a rescaled version of $\Sa$. Following \citep{chen2018neural} we use the \textit{number of function evaluations} (NFE) of the numerical solver utilized to solve (\ref{eq:GDE}) as a proxy for model depth. In applications where $\Sa$ acquires a specific meaning (i.e forecasting with irregular timestamps) the integration domain can be appropriately tuned to evolve GDE dynamics between \textit{arrival times} \citep{rubanova2019latent} without assumptions on the functional form of the underlying vector field, as is the case for example with exponential decay in GRU--D \citep{che2018recurrent}.
\paragraph{GDE training} GDEs can be trained with a variety of methods. Standard backpropagation through the computational graph, adjoint sensitivity method \citep{pontryagin1962mathematical} for $\mathcal{O}(1)$ memory efficiency \citep{chen2018neural}, or backpropagation through a relaxed spectral elements discretization \citep{quaglino2019accelerating}. 
Numerical instability in the form of accumulating errors on the adjoint ODE during the backward pass of Neural ODEs has been observed in \citep{gholami2019anode}. A proposed solution is a hybrid checkpointing--adjoint scheme commonly employed in scientific computing \citep{wang2009minimal}, where the adjoint trajectory is reset at predetermined points in order control the error dynamics.
\section{Taxonomy of GDEs}
In the following, we taxonomize GDEs models distinguishing them into \textit{static} and \textit{spatio--temporal} (autoregressive) variants. 

\subsection{Static Models}
\paragraph{Graph convolution differential equations} Based on graph spectral theory \citep{shuman2013emerging,sandryhaila2013discrete}, the residual version of \textit{graph convolution network} (GCN) \citep{kipf2016semi} layers are in the form:
\begin{equation}\label{eq:gcn}
    \Hb{(s+1)} = \Hb(s) + \sigma\left(\Lb_{\G} \Hb(s) \parM(s)\right)
\end{equation}
where $\Lb_{\G}\in\R^{n\times n}$ is the graph \textit{Laplacian} and $\sigma$ is as nonlinear activation function. We denote with $\C_{\G}$ the graph convolution operator, i.e.
$\C_{\G}\Hb(s)= \Lb_{\G} \Hb(s) \parM(s)$. A general formulation of the continuous counterpart of GCNs, \textit{graph convolution differential equation} (GCDE), is therefore obtained by defining $\Fb_{\G}$ as a multilayer convolution, i.e.
\begin{equation}\label{eq:gde}
    \dot\Hb{(s)} = \Fb_{\tt GCN}(\Hb(s), \parM) := \C^N_{\G}\circ ~\sigma\circ\C^{N-1}_{\G}\circ\cdots\circ\sigma\circ\C^1_{\G}\Hb(s)
\end{equation}
Note that the Laplacian $\Lb_{\G}$ can be computed in different ways, see e.g. \citep{bruna2013spectral,defferrard2016convolutional,levie2018cayleynets, zhuang2018dual}. Diffusion--type convolution layers \citep{li2017diffusion} are also compatible with the continuous--depth formulation.
\paragraph{Additional models and considerations} We include additional derivation of continuous counterparts of common static GNN models such as \textit{graph attention networks} (GAT) \citep{velivckovic2017graph} and general message passing GNNs as supplementary material. 
\subsection{Spatio--Temporal Models}
For settings involving a temporal component (i.e., modeling dynamical systems), the depth domain of GDEs coincides with the time domain $s\equiv t$ and can be adapted depending on the requirements. For example, given a time window $\Delta t$, the prediction performed by a GDE assumes the form:
\begin{equation*}
    \Hb{(t + \Delta t)} = \Hb(t) + \int_t^{t + \Delta t}\Fb\lc \tau,\Hb(\tau),\parM \rc d\tau,
\end{equation*}
regardless of the specific GDE architecture employed. Here, GDEs represent a natural model class for autoregressive modeling of sequences of graphs $\{\G_t\}$ and seamlessly link to dynamical network theory. This line of reasoning naturally leads to an extension of classical spatio--temporal architectures in the form of \textit{hybrid dynamical systems} \citep{van2000introduction,goebel2008}, i.e., systems characterized by interacting continuous and discrete--time dynamics.
 Let  $(\K, >)$, $(\T, >)$ be linearly ordered sets; namely,  $\K\subset\Nat\setminus\{0\}$ and $\T$ is a set of time instants, $\T:=\{t_k\}_{k\in\K}$. We suppose to be given a \textit{state--graph data stream} which is a sequence in the form $\left\{\left(\Xb_t,\G_t\right)\right\}_{t\in\T}$. 
Let us also define a \textit{hybrid time domain} as the set $\I:= \bigcup_{k\in\K}\lc[t_k, t_{k+1}],k\rc$ and a \textit{hybrid arc} on $\I$ as a function $\bm\Phi$ such that for each $k\in\K$, $t\mapsto\bm\Phi(t,k)$ is absolutely continuous in $\{t:(t,j)\in\dom\bm\Phi\}$. Our aim is to build a continuous model predicting, at each $t_k\in\T$, the value of $\Xb_{t_{k+1}}$, given $\left(\Xb_t,\G_t\right)$.
The core idea is to have a GDE smoothly steering the latent node features between two time instants and then apply some discrete operator, resulting in a ``jump'' of $\Hb$ which is then processed by an output layer. Solutions of the proposed continuous spatio--temporal model are therefore hybrid arcs.  
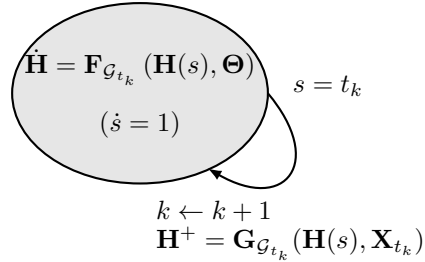
\begin{wrapfigure}[10]{r}{0.4\textwidth}
	\centering
	\begin{tikzpicture}%[scale=.85]
	\fill[gray!20, draw = black,thick] (0,0) ellipse (1.7cm and 1.2cm);
	\draw (0,0) node[align = center](flow)  {$\dot{\Hb} = \Fb_{\G_{t_k}}\left(\Hb(s),\parM\right)$\\\\$(\dot s= 1)$};
	\draw [thick,-latex] (1.7,0) to [in=-30, out=-50, distance=1.2cm] (.9,-1.);

	\draw (2,-1.75) node[align = left] {\\{$k\leftarrow k+1$}\\
 	$\Hb^+ = \mathbf{G}_{\G_{t_k}}(\Hb(s),\Xb_{t_{k}})$};
	\draw (2.5,0.1) node {$s={t_{k}}$};
	\end{tikzpicture}
	\vspace{-7mm}
	\caption{Schematic of autoregressive GDEs as hybrid automata.}
	\label{fig:aut}
\end{wrapfigure}
\paragraph{Autoregressive GDEs} 
The solution of a general autoregressive GDE model can be symbolically represented by:
\begin{equation}\label{eq:hybrid}
   \left\{
        \begin{matrix*}[l]
            \dot{\Hb}({s}) &= \Fb_{\G_{t_k}}(\Hb(s), \parM) & s\in[t_{k-1}, t_{k}] \\[3pt]

            \Hb^+(s) &= \mathbf{G}_{\G_{t_k}}(\Hb(s), \Xb_{t_k}) & s = t_{k}\\[3pt]
           \Yb &= \mathbf{K}(\Hb(s)) & s=t_{k}
        \end{matrix*}
    \right.k\in\K,
\end{equation} 
where $\Fb, \mathbf{G}, \mathbf{K}$ are GNN--like operators or general neural network layers\footnote{More formal definitions of the hybrid model in the form of \textit{hybrid inclusions} can indeed be easily given. However, the technicalities involved are beyond the scope of this paper.} and $\Hb^+$ represents the value of $\Hb$ after the discrete transition. The evolution of system (\ref{eq:hybrid}) is indeed a sequence of hybrid arcs defined on a hybrid time domain. A graphical representation of the overall system is given by the \textit{hybrid automata} as shown in Fig. \ref{fig:aut}. Compared to standard recurrent models which are only equipped with discrete jumps, system (\ref{eq:hybrid}) incorporates a continuous flow of latent node features $\Hb$ between jumps. This feature of autoregressive GDEs allows them to track the evolution of dynamical systems from observations with irregular time steps.
In the experiments we consider $\mathbf{G}$ to be a GRU cell \citep{cho2014learning}, obtaining \textit{graph convolutional differential equation--GRU}  (GCDE--GRU). Alternatives such as GCDE--RNNs or GCDE--LSTMs can be similarly obtained by replacing $\mathbf{G}$ with other common recurrent modules, such as vanilla RNNs or LSTMs \citep{hochreiter1997long}. It should be noted that the operators $\Fb, \mathbf{G}, \mathbf{K}$ can themselves have multi--layer structure. 
\section{Experiments}
We evaluate GDEs on a suite of different tasks. The experiments and their primary objectives are summarized below:
\begin{itemize}
  \item Semi--supervised node classification on static, standard benchmark datasets Cora, Citeseer, Pubmed \citep{sen2008collective}. We investigate the usefulness of the proposed method in a static setting via an ablation analysis that directly compares GCNs and analogue GCDEs solved with fixed--step and adaptive solvers.
    \item Trajectory extrapolation task on a synthetic multi--agent dynamical system. We compare Neural ODEs and GDEs, providing a motivating example for the introduction of additional biases in the form of second--order models \citep{yildiz2019ode,massaroli2020dissecting}. 
    \item Traffic forecasting on an undersampled version of PeMS \citep{yu2018spatio} dataset. We measure the performance improvement obtained by a correct inductive bias on continuous dynamics and robustness to irregular timestamps.
\end{itemize}
The code will be open--sourced after the review phase and is included in the submission.
\begin{figure*}[t]
    \centering
    \includegraphics[width=1\linewidth]{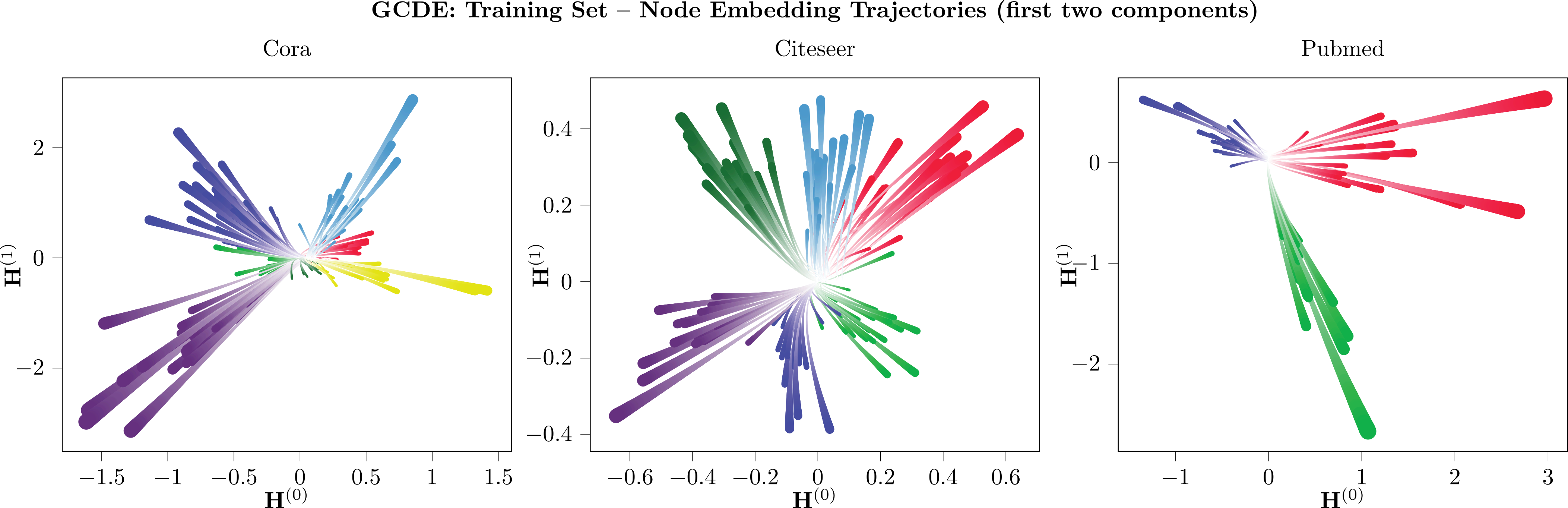}
    \caption{Node embedding trajectories defined by a forward pass of GCDE--dpr5 on Cora, Citeseer and Pubmed. Color differentiates between node classes.}
    \label{fig:node_emb_traj}
\end{figure*}
\begin{table}[b]
\footnotesize
\centering
\setlength\tabcolsep{4pt}
\begin{tabular}{lrrr}
\toprule
Model (NFE) & Cora & Citeseer & Pubmed\\
\midrule
GCN & $81.4 \pm 0.5\%$ & $70.9 \pm 0.5\%$ & $79.0 \pm 0.3\%$\\
GCN$^*$ & $82.8 \pm 0.3\%$ & $71.2 \pm 0.4\%$ & $79.5 \pm 0.4\%$ \\
\midrule
GCDE--rk2 (2) & $83.0 \pm 0.6\%$ &  $72.3 \pm 0.5\%$ & $\textbf{79.9} \pm 0.3\%$\\
GCDE--rk4 (4) & $\textbf{83.8} \pm 0.5\%$ &  $\textbf{72.5} \pm 0.5\%$ & $79.5 \pm 0.4\%$\\
GCDE--dpr5 \textbf{(158)} & $81.8 \pm 1.2\%$ & $68.3\pm 1.2\%$ & $78.5 \pm 0.7\%$\\
\bottomrule
\end{tabular}
\vspace{3mm}
\caption{Test results across 100 runs ($\mu$ and $\sigma$). All models have hidden dimension $64$.}
\label{tab:allresone}
\end{table}
\subsection{Transductive Node Classification}
\paragraph{Experimental setup}
The first task involves performing semi--supervised node classification on static graphs collected from baseline datasets Cora, Pubmed and Citeseer \citep{sen2008collective}. Main goal of these experiments is to perform an ablation study on the source of possible performance advantages of the GDE framework in settings that do not involve continuous dynamical systems. 
The $L_2$ weight penalty is set to $5\cdot 10^{-4}$ on Cora, Citeseer and $10^{-3}$ on Pubmed as a strong regularizer due to the small size of the training set \citep{monti2017geometric}.  We report mean and standard deviation across $100$ training runs. Since our experimental setup follows \citep{kipf2016semi} to allow for a fair comparison, other baselines present in recent GNN literature can be directly compared with Table \ref{tab:allresone}.
\paragraph{Models and baselines}
All convolution--based models are equipped with a latent dimension of $64$. We include results for best performing vanilla GCN baseline presented in \citep{velivckovic2017graph}. To avoid flawed comparisons, we further evaluate an optimized version of GCN, GCN$^*$, sharing exact architecture, as well as training and validation hyperparameters with the GCDE models. We experimented with different number of layers for GCN$^*$: (2, 3, 4, 5, 6) and select $2$, since it achieves the best results. The performance of \textit{graph convolution differential equation} (GCDE) is assessed with both a fixed-step solver Runge--Kutta \citep{runge1895numerische,kutta1901beitrag} as well as an adaptive--step solver, Dormand--Prince \citep{dormand1980family}. The resulting models are denoted as GCDE--rk4 and  GCDE--dpr5, respectively. We utilize the {\tt torchdiffeq} \citep{chen2018neural} PyTorch package to solve and backpropagate through the ODE solver.
\paragraph{Continuous--depth models in static tasks}
Ensuring a low error solution to the ODE parametrized by the model with adaptive--step solvers does not offer particular advantages in image classification tasks \citep{chen2018neural} compared to equivalent discrete models. While there is no reason to expect performance improvements solely from the transition away from discrete architectures, continuous--depth allows for the embedding of numerical ODE solvers in the forward pass. Multi--step architectures have previously been linked to ODE solver schemes \citep{lu2017beyond} and routinely outperform their single--step counterparts \citep{larsson2016fractalnet,lu2017beyond}. We investigate the performance gains by employing the GDE framework in static settings as a straightforward approach to the embedding of numerical ODE solvers in GNNs. 
\paragraph{Results}
The variants of GCDEs solved with fixed--step schemes are shown to outperform or match GCN$^*$ across all datasets, with the margin of improvement being highest on Cora and Citeseer. Introducing GCDE--rk2 and GCDE--rk4 is observed to provide the most significant accuracy increases in more densely connected graphs or with larger training sets. In particular, GCDE--rk4 outperforming GCDE--rk2 indicates that, given equivalent network architectures, higher order ODE solvers are generally more effective, provided the graph is dense enough to benefit from the additional computation. Additionally, training GCDEs with adaptive step solvers naturally leads to deeper models than possible with vanilla GCNs, whose layer depth greatly reduces performance. However, the high \textit{number of function evaluation} (NFEs) of GCDE--dpr5 necessary to stay within the ODE solver tolerances causes the model to overfit and therefore generalize poorly. We visualize the first two components of GCDE--dpr5 node embedding trajectories in Figure \ref{fig:node_emb_traj}. The trajectories are divergent, suggesting a non--decreasing classification performance for GCDE models trained with longer integration times. We provide complete visualization of accuracy curves in Appendix C. 
\begin{wrapfigure}[10]{r}{0.42\textwidth}
    \vspace{-5mm}
    \centering
     \includegraphics[width=1\linewidth]{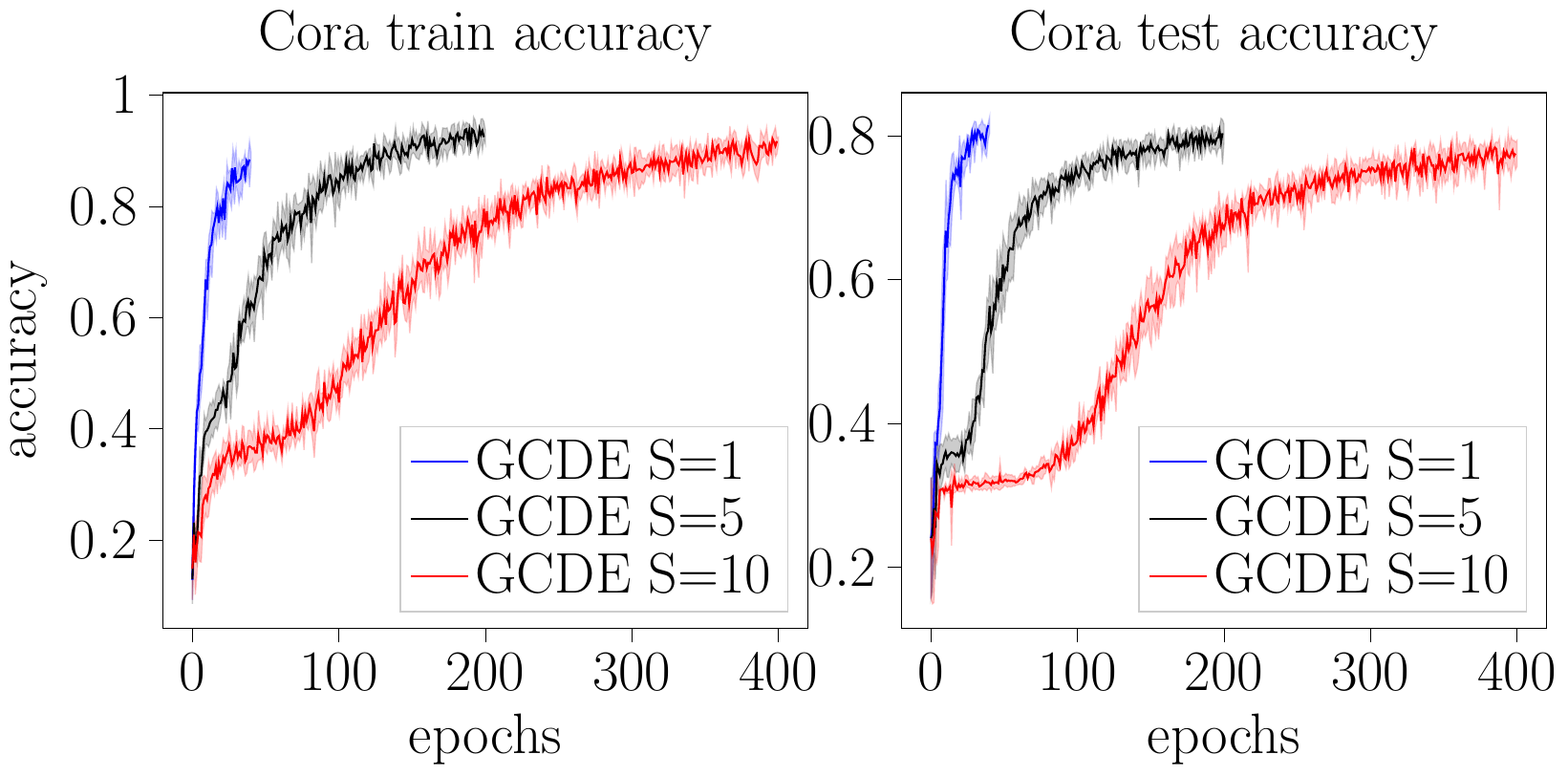}
    \caption{Cora accuracy of GCDE models with different integration times $s$.}
    \label{fig:trk4}
\end{wrapfigure}
\paragraph{Resilience to integration time}
For each integration time $S \in [1, 5, 10]$, we train $100$ GCDE-dpr5 models on Cora and report average metrics, along with $1$ standard deviation confidence intervals in Figure \ref{fig:trk4}. GCDEs are shown to be resilient to changes in $S$; however, GCDEs with longer integration times require more training epochs to achieve comparable accuracy. This result suggests that, indeed, GDEs are immune to node oversmoothing \citep{oono2019graph}.
\subsection{Multi--Agent Trajectory Extrapolation}
\paragraph{Experimental setup}
We evaluate GDEs and a collection of deep learning baselines on the task of extrapolating the dynamical behavior of a synthetic mechanical multi--particle system. Particles interact within a certain radius with a viscoelastic force. Outside the mutual interactions, captured by a time--varying adjacency matrix $\Ab_t$, the particles would follow a periodic motion. The adjaciency matrix $\Ab_t$ is computed along the trajectory as:
\[
    \Ab_t^{(ij)} = 
    \left\{\begin{matrix*}[l]
        1 & 2\|\x_i(t)-\x_j(t)\|\leq r\\
        0 & \text{otherwise}
    \end{matrix*}\right.~,
\]
\begin{wrapfigure}[11]{r}{0.45\textwidth}
    \vspace{-5mm}
    \centering
     \includegraphics[width=1\linewidth]{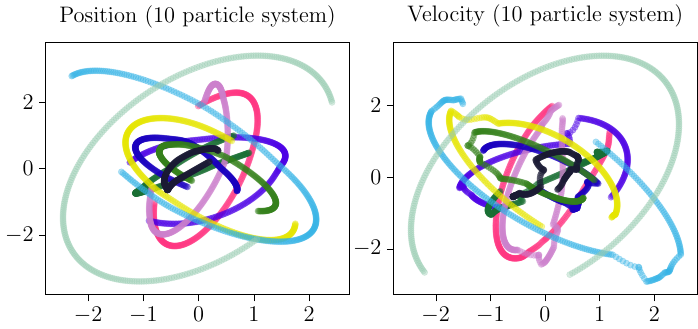}
    \caption{Example position and velocity trajectories of the multi--particle system.}
    \label{fig:ma_traj}
\end{wrapfigure}
where $\x_i(t)$ is the position of node $i$ at time $t$. Therefore, $\Ab_t$ results to be symmetric, $\Ab_t = \Ab_t^\top$ and yields an undirected graph. The dataset is collected by integrating the system for $T = 5s$ with a fixed step--size of $dt = 1.95\cdot10^{-3}$ and is split evenly into a training and test set. We consider $10$ particle systems. An example trajectory is shown in Figure \ref{fig:ma_traj}. All models are optimized to minimize mean--squared--error (MSE) of 1--step predictions using Adam \citep{kingma2014adam} with constant learning rate $0.01$. 
We measure test \textit{mean average percentage error} (MAPE) of model predictions in different extrapolation settings. \textit{Extrapolation steps} denotes the number of predictions each model $\Phi$ has to perform without access to the nominal trajectory. This is achieved by recursively letting inputs at time $t$ be model predictions at time $t - \Delta t$ i.e $\hat{\Yb}_{t+\Delta t} = \phi (\hat{\Yb}_{t})$ for a certain number of extrapolation steps, after which the model is fed the actual nominal state $\Xb$ and the cycle is repeated until the end of the test trajectory. For a robust comparison, we report mean and standard deviation across 10 seeded training and evaluation runs. Additional experimental details, including the analytical formulation of the dynamical system, are provided as supplementary material. 
\paragraph{Models and baselines}
As the vector field depends only on the state of the system, available in full during training, the baselines do not include recurrent modules. We consider the following models:
\begin{itemize}
    \item A 3--layer fully-connected neural network, referred to as \textit{Static}. No assumption on the dynamics
    \item A vanilla Neural ODE with the vector field parametrized by the same architecture as \textit{Static}. ODE assumption on the dynamics.
    \item A 3--layer convolution GDE, GCDE. Dynamics assumed to be determined by a blend of graphs and ODEs
    \item A 3--layer, second--order \citep{yildiz2019ode,massaroli2020dissecting} GCDE and referred to as \textit{GCDE-II}. GCDE assumptions in addition to second--order ODE dynamics.
\end{itemize}
A grid hyperparameter search on number of layers, ODE solver tolerances and learning rate is performed to optimize \textit{Static} and Neural ODEs. We use the same hyperparameters for GDEs. 
\begin{wrapfigure}[13]{r}{0.45\textwidth}
\vspace{-5mm}
    \centering
    \includegraphics[width=1\linewidth]{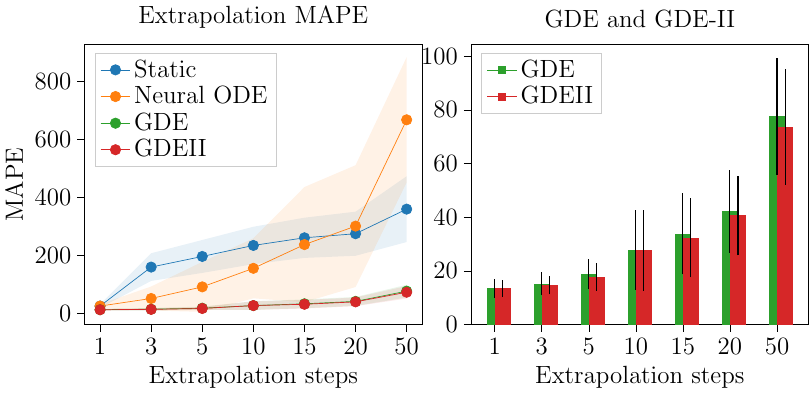}
    \caption{Test extrapolation MAPE averaged across 10 experiments. Shaded area and error bars indicate 1--standard deviation intervals. }
    \label{fig:mape}
\end{wrapfigure}
\paragraph{Results}
Figure \ref{fig:mape} shows the growth rate of test MAPE error as the number of extrapolation steps is increased. \textit{Static} fails to extrapolate beyond the 1--step setting seen during training. Neural ODEs overfit spurious particle interaction terms and their error rapidly grows as the number of extrapolation steps is increased. GCDEs, on the other hand, are able to effectively leverage relational information to track the system: we provide complete visualization of extrapolation trajectory comparisons in the supplementary material. Lastly, GCDE-IIs outperform first--order GCDEs as their structure inherently possesses crucial information about the relative relationship of positions and velocities that is accurate with respect to the observed dynamical system. 
\subsection{Traffic Forecasting}
\paragraph{Experimental setup}
We evaluate the effectiveness of autoregressive GDE models on forecasting tasks by performing a series of experiments on the established PeMS traffic dataset. We follow the setup of \citep{yu2018spatio} in which a subsampled version of PeMS, PeMS7(M), is obtained via selection of 228 sensor stations and aggregation of their historical speed data into regular 5 minute frequency time series. We construct the adjacency matrix $\Ab$ by thresholding of the Euclidean distance between observation stations i.e. when two stations are closer than the threshold distance, an edge between them is included. The threshold is set to the 40$^{\text{th}}$ percentile of the station distances distribution. To simulate a challenging environment with missing data and irregular timestamps, we undersample the time series by performing independent Bernoulli trials on each data point. Results for 3 increasingly challenging experimental setups are provided: undersampling with $30\%$, $50\%$ and $70\%$ of removal. In order to provide a robust evaluation of model performance in regimes with irregular data, the testing is repeated $20$ times per model, each with a different undersampled version of the test dataset. We collect \textit{root mean square error} (RMSE) and MAPE. More details about the chosen metrics and data are included as supplementary material.
\paragraph{Models and baselines}
In order to measure performance gains obtained by GDEs in settings with data generated by continuous time systems, we employ a GCDE--GRU--dopri5 as well as its discrete counterpart GCGRU \citep{zhao2018deep}. To contextualize the effectiveness of introducing graph representations, we include the performance of GRUs since they do not directly utilize structural information of the system in predicting outputs. Apart from GCDE--GRU, both baseline models have no innate mechanism for handling timestamp information. For a fair comparison, we include timestamp differences between consecutive samples and \textit{sine--encoded} \citep{petnehazi2019recurrent} absolute time information as additional features. All models receive an input sequence of $5$ graphs to perform the prediction.
 \begin{table*}[b]
\tiny
\centering
\setlength\tabcolsep{3pt}
\begin{tabular}{lrrrrrrrr}
\toprule
Model & MAPE$_{30\%}$ & RMSE$_{30\%}$ & MAPE$_{50\%}$ & RMSE$_{50\%}$ & MAPE$_{70\%}$ & RMSE$_{70\%}$ &  MAPE$_{100\%}$ & RMSE$_{100\%}$\\
\midrule
GRU & $27.14 \pm 0.45$ & $13.25 \pm 0.11$ & $27.08 \pm 0.26$ & $13.22 \pm 0.07$ & $27.24 \pm 0.19$ & $13.28 \pm 0.05$ & $27.20$ & $13.29$\\
GCGRU & $23.60 \pm 0.38$ & $11.97 \pm 0.03$ & $22.86 \pm 0.22$ & $11.78 \pm 0.06$ & $21.33 \pm 0.16$ & $11.20 \pm 0.04$ &$20.92$ & $10.87$\\
GCDE-GRU & $\mathbf{22.95} \pm 0.37$ & $\mathbf{11.67} \pm 0.10$ & $\mathbf{21.25} \pm 0.21$ & $\mathbf{11.04} \pm 0.05$ & $\mathbf{20.94} \pm 0.14$ & $\mathbf{10.95} \pm 0.04$ & $\mathbf{20.46}$ & $\mathbf{10.766}$\\
\bottomrule
\end{tabular}
\caption{Forecasting test results across 20 runs (mean and standard deviation). MAPE$_i$ indicates $i\%$ undersampling of the test set.}
\label{tab:traffic_preds}
\end{table*}
\paragraph{Results}
Non--constant differences between timestamps result in a challenging forecasting task for a single model since the average prediction horizon changes drastically over the course of training and testing. 
Traffic systems are intrinsically dynamic and continuous in nature and, therefore, a model able to track continuous underlying dynamics is expected to offer improved performance.
Since GCDE-GRUs and GCGRUs are designed to match in structure we can measure this performance increase from the results shown in Table \ref{tab:traffic_preds}. GCDE--GRUs outperform GCGRUs and GRUs in all undersampling regimes. Additional details and prediction visualizations are included in Appendix C.
\section{Related work}
There exists a concurrent line of work \citep{xhonneux2019continuous} introducing a GNN variant evaluated on static node classification tasks where the output is the analytical solution of a linear ODE. \cite{sanchez2019hamiltonian} proposes using graph networks (GNs) \citep{battaglia2018relational} and ODEs to track Hamiltonian functions, whereas \citep{deng2019continuous} introduces a GNN version of continuous normalizing flows \citep{chen2018neural,grathwohl2018ffjord} for generative modeling, extending \citep{liu2019graph}. Our goal is developing a unified system--theoretic framework for continuous--depth GNNs covering the main variants of static and spatio--temporal GNN models. Our work provides extensive experimental evaluations on both static as well as dynamic tasks with the primary aim of uncovering the sources of performance improvement of GDEs in each setting.
\section{Discussion}
\paragraph{Unknown or dynamic topology} Several lines of work concerned with learning the graph structure directly from data exist, either by inferring the adjacency matrix within a probabilistic framework \citep{kipf2018neural} or using a soft--attention \citep{vaswani2017attention} mechanism \citep{choi2017gram,li2018adaptive,wu2019graph}. In particular, the latter represents a commonly employed approach to the estimation of a dynamic adjacency matrix in spatio--temporal settings. Due to the algebraic nature of the relation between the attention operator and the node features, GDEs are compatible with its use inside the GNN layers parametrizing the vector field. Thus, if an optimal adaptive graph representation $\mathbf{S}(s, \Hb)$ is computed through some attentive mechanism, standard convolution GDEs can be replaced by $\dot\Hb = \sigma\lc\mathbf{S}\Hb\parM\rc.$
\paragraph{Addition or removal of nodes}
GDE variants operating on sequences of dynamically changing graphs can, without changes to the formulation, directly accommodate addition or removal of nodes as long as its number remains constant during the flows. In fact, the size of parameter matrix $\bm\Theta$ exclusively depends on the node feature dimension, resulting in resilience to a varying number of nodes.
\section{Conclusion}
In this work we introduce \textit{graph neural ordinary differential equations} (GDE), the continuous--depth counterpart to \textit{graph neural networks} (GNN) where the inputs are propagated through a continuum of GNN layers. The GDE formulation is general, as it can be adapted to include many static and autoregressive GNN models. GDEs are designed to offer a data--driven modeling approach for \textit{dynamical networks}, whose dynamics are defined by a blend of discrete topological structures and differential equations. In sequential forecasting problems, GDEs can accommodate irregular timestamps and track the underlying continuous dynamics, whereas in static settings they offer computational advantages by allowing for the embedding of black--box numerical solvers in their forward pass.  GDEs have been evaluated on both static and dynamic tasks and have been shown to outperform their discrete counterparts. 
\newpage
%
% \section*{Broader Impact}
% Multi--agent systems permeate science in a variety of fields: from physics to robotics, game-theory, finance and molecular biology, among others. Based on the interplay between nonlinear dynamical systems and graphs, \textit{dynamical network} theory has been developed as a widely applicable, classical approach to control and stabilization of such systems. The primary purpose of GDEs is to offer a data--driven approach to the modeling of dynamical networks, particularly when the governing equations are highly nonlinear and therefore challenging to approach with classical or analytical methods. Depending on the speed of adoption of these novel continuous--depth models, fields such as healthcare which are already seeing preliminary attempts \citep{rubanova2019latent,yildiz2019ode}, will have access to potentially more accurate forecasting models. Outside of positive impacts on patients and customers of such technology, this would ultimately lead to ripple effects on employment as job availability is affected by automation.  

\bibliographystyle{abbrvnat}
\bibliography{main.bib}
\newpage
\rule[0pt]{\columnwidth}{3pt}
\begin{center}
\huge{\bf{Graph Neural Ordinary Differential Equations} \\
\emph{Supplementary Material}}
\end{center}
\vspace*{3mm}
\rule[0pt]{\columnwidth}{1pt}%\hline
\vspace*{-.5in}
\appendix
\addcontentsline{toc}{section}{}
\part{}
\parttoc
\section{Graph Neural Ordinary Differential Equations}
\paragraph{Notation}
Let $\Nat$ be the set of natural numbers and $\R$ the set of reals. Scalars are indicated as lowercase letters, vectors as bold lowercase, matrices and tensors as bold uppercase and sets with calligraphic letters. Indices of arrays and matrices are reported as superscripts in round brackets.

Let $\V$ be a finite set with $|\V| = n$ whose element are called \textit{nodes} and let $\E$ be a finite set of tuples of $\V$ elements. Its elements are called \textit{edges} and are such that $\forall e_{ij}\in\E,~e_{ij} = (v_i,v_j)$ and $v_i,v_j\in\V$. A graph $\G$ is defined as the collection of nodes and edges, i.e. $\G := (\V,\E)$. The \textit{adjeciency} matrix $\Ab\in\R^{n\times n}$ of a graph is defined as
\[
    \Ab^{(ij)} = \left\{
        \begin{matrix*}[l]
        1 & e_{ij}\in\E\\
        0 & e_{ij}\not\in\E
        \end{matrix*}
        \right.~.
\]
If $\G$ is an \textit{attributed graph}, the \textit{feature vector} of each $v\in\V$ is $\x_v\in\R^{n_x}$. All the feature vectors are collected in a matrix $\Xb\in\R^{n\times n_x}$. Note that often, the features of graphs exhibit temporal dependency, i.e. $\Xb := \Xb_t$.
\clearpage
\subsection{General Static Formulation}
For clarity and as an easily accessible reference, we include below a general formulation table for the static case
\begin{CatchyBox}{Graph Neural Ordinary Differential Equations}
    \begin{minipage}[h]{0.45\linewidth}%\small
            \begin{equation*}\label{eq:1}%\boxed
            %\tcboxmath
            {
                \left\{
                \begin{aligned}
                    \dot\Hb(s) &= \Fb_{\G}\lc s, \Hb(s), \parM\rc\\
                    \Hb(0) &= \Xb_e\\
                    \Yb(s) &= \mathbf{K}(\Hb(s))
                \end{aligned}
                \right.~~
                s\in S}
            \end{equation*}
            \hfill
    \end{minipage}
    \hfill
    \begin{minipage}[h]{.51\linewidth}\small
        \centering
        \begin{tabular}{r|c|c}
            Input & $\Xb$ & $\R^{n\times n_x}$\\\hline
            Embedded Input & $\Xb_e$ & $\R^{n\times h}$\\\hline
            Output & $\Yb(s)$ & $\R^{n\times n_y}$\\\hline
            Graph & $\G$ & \\\hline
            Node features & $\Hb$ & $\R^{n\times h}$\\\hline
            Parameters & $\parM$ & $\R^{h\times h}$\\\hline
            Neural Vector Field & $\Fb_{\G}$ & $\R^h \rightarrow \R^h$\\\hline
            Output Network & $\mathbf{K}$ & $\R^h \rightarrow \R^{n_y}$\\
        \end{tabular}
    \end{minipage}
\end{CatchyBox}
Note that the general formulation provided in \eqref{eq:hybrid} can similarly serve as a reference for the spatio--temporal case.
\subsection{Computational Overhead}
As is the case for other models sharing the continuous--depth formulation \citep{chen2018neural}, the computational overhead required by GDEs depends mainly by the numerical methods utilized to solve the differential equations. We can define two general cases for \textit{fixed--step} and \textit{adaptive--step} solvers.
\paragraph{Fixed--step}
In the case of fixed--step solvers of \textit{k--th} order e.g \textit{Runge--Kutta--k} \citep{runge1895numerische}, the time complexity is $O(nk)$ where $n := S/\epsilon$ defines the number of steps necessary to cover $[0, S]$ in fixed--steps of $\epsilon$.
\paragraph{Adaptive--step}
For general adaptive--step solvers, computational overhead ultimately depends on the error tolerances. While worst--case computation is not bounded \citep{dormand1980family}, a maximum number of steps can usually be set algorithmically. 
\subsection{Additional GDEs}
\paragraph{Message passing GDEs}  Let us consider a single node $v\in\V$ and define the set of neighbors of $v$ as $\N(v): = \{u\in\V~:~(v,u)\in\E \lor (u,v)\in\E\}$. Message passing neural networks (MPNNs) perform a spatial--based convolution on the node $v$ as
\begin{equation}\label{eq:MPNN}
    \h^{(v)}{(s+1)} = \ub\left[\h^{(v)}(s), \sum_{u\in\N(v)}\m\lc{\h^{(v)}(s),\h^{(u)}(s)}\rc\right],
\end{equation}
where, in general, $\h^{v}(0) = \x_v$ while $\ub$ and $\m$ are functions with trainable parameters. For clarity of exposition, let $\ub(\x,\y) := \x + \g(\y)$ where $\g$ is the actual parametrized function. The (\ref{eq:MPNN}) becomes
\begin{equation}
    \h^{(v)}{(s+1)} = \h^{(v)}(s) + \g\left[\sum_{u\in\N(v)}\m\lc{\h^{(v)}(s),\h^{(u)}(s)}\rc\right],
\end{equation}
and its continuous--depth counterpart, \textit{graph message passing differential equation} (GMDE) is:
\begin{equation*}
    \dot\h^{(v)}{(s)} = \f^{(v)}_{\tt MPNN}(\Hb, \parM) := \g\left[\sum_{u\in\N(v)}\m\lc{\h^{(v)}(s),\h^{(u)}}(s)\rc\right].
\end{equation*}
%
% \paragraph{Diffusion graph convolution}
% Consider a diffusion graph convolution (DGC) network \citep{atwood2016diffusion}:
% %
% \begin{equation}\label{eq:DCG}
%     \Hb_{s+1} = \Hb_s + \sigma\lc\Pb^s\Xb\Wb_s\rc
% \end{equation}
% %
% where $\sigma:\R\rightarrow\R$ is an activation function assumed to act component--wise, $\Pb\in\R^{n\times n}$ is a probability transition matrix, $\Pb:=\Db^{-1}\Ab$.
% The continuous counterpart of (\ref{eq:DCG}), DGC--NODE, can be therefore derived as:
% %
% \begin{equation*}
%     \frac{d\Hb}{d s} = \Fb_{\tt DGC}(s, \Xb, \parM) := \sigma\lc\Pb^s\Xb\parM\rc,\quad
% \end{equation*}
% %
% which consists in a depth--varying vector field constant with respect to the hidden node feature matrix.
%
\paragraph{Attention GDEs} Graph attention networks (GATs) \citep{velivckovic2017graph} perform convolution on the node $v$ as
\begin{equation}
    \h^{(v)}{(s+1)} = \sigma\lc\sum_{u\in\N(v)\cup v}{\alpha_{vu}\parM(s)\h^{(u)}}(s)\rc.
\end{equation}
Similarly, to GCNs, a \textit{virtual} skip connection can be introduced allowing us to define the \textit{graph attention differential equation} (GADE):
\begin{equation*}
    \dot\h^{(v)}{(s)} = \f^{(v)}_{\tt GAT}(\Hb,\parM):= \sigma\lc\sum_{u\in\N(v)\cup v}{\alpha_{vu}\parM\h^{(u)}}(s)\rc,
\end{equation*}
where $\alpha_{vu}$ are attention coefficient which can be computed following \citep{velivckovic2017graph}.
\section{Spatio--Temporal GDEs}
We include a complete description of GCGRUs to clarify the model used in our experiments.
%\subsection{Note on the naming conventions}
%I use two different types of notations for describing two general types of GNN and RNN connections.
%\begin{itemize}
%    \item \textit{GNN}-\textit{RNN}: The models where uses GNN to encode graph features and use the output of GNN module (or \textit{read-out} of it) to the input of the RNN module. e.g) GC-GRU
%    \item \textit{GNN}\textit{RNN}: The models whose matrix multiplications of RNN modules are replaced with GNN e.g) GCGRU
%\end{itemize}
%
\begin{figure*}[t]
    \centering
    \includegraphics[width=1\linewidth]{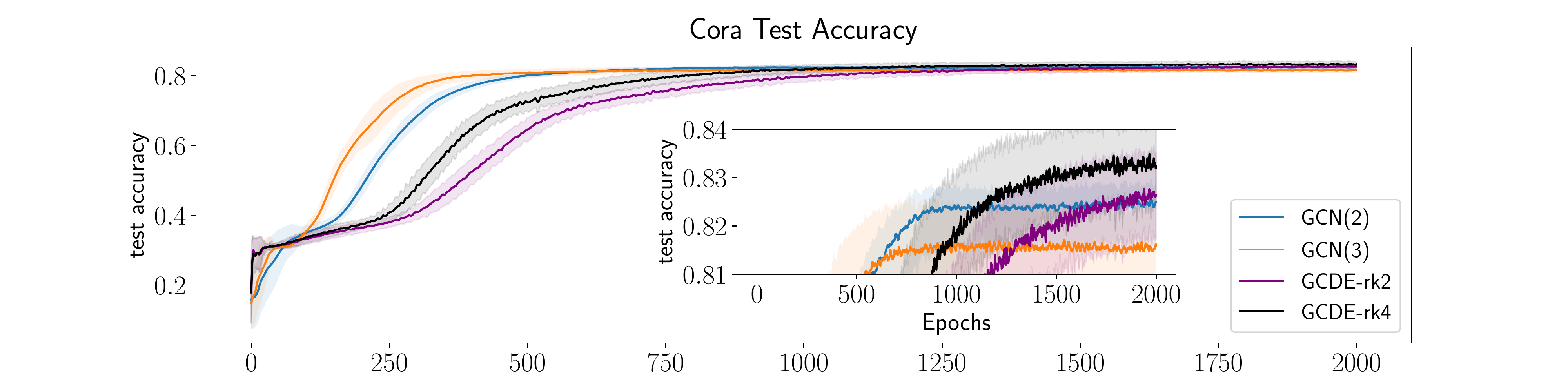}
    \includegraphics[width=1\linewidth]{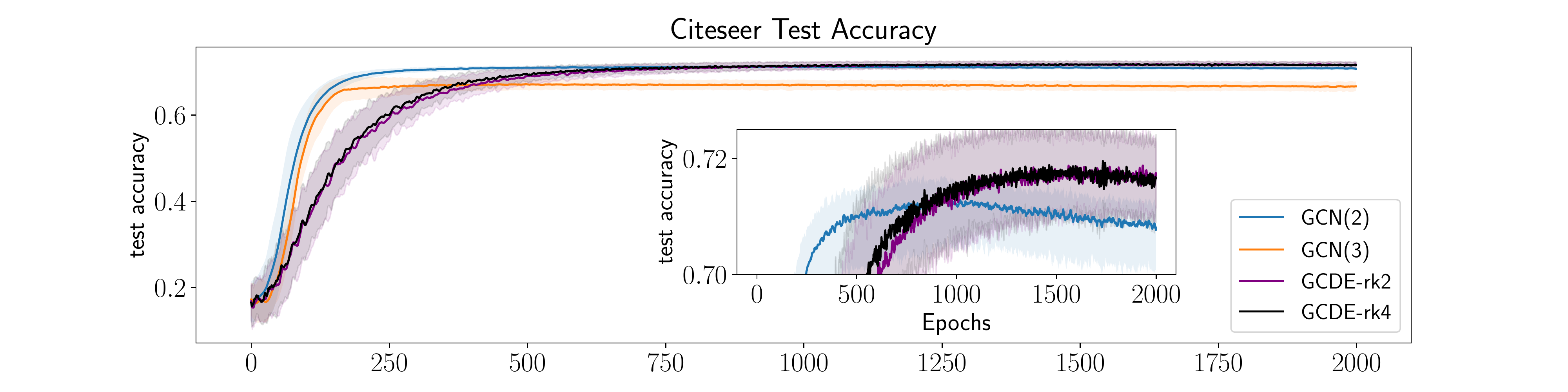}
    \caption{Test accuracy curves on Cora and Citeseer (100 experiments). Shaded area indicates the 1 standard deviation interval.}
    \label{fig:curves}
\end{figure*}
\subsection{GCGRU Cell}
Following GCGRUs \citep{zhao2018deep}, we perform an instantaneous jump of $\Hb$ at each time $t_{k}$ using the next input features $\Xb_{t_{k}}$. Let $\Lb_{\G_{t_k}}$ be the graph Laplacian of graph $\G_{t_k}$, which can computed in several ways \citep{bruna2013spectral,defferrard2016convolutional,levie2018cayleynets, zhuang2018dual}. Then, let
\begin{equation}
\begin{aligned} 
\mathbf{Z} &:=\sigma\left(\Lb_{\G_{t_k}}\Xb_{t_k}\parM_{xz} + \Lb_{\G_{t_k}}\Hb\parM_{hz}\right), \\
\mathbf{R} &:=\sigma\left(\Lb_{\G_{t_k}}\Xb_{t_k}\parM_{xr} + \Lb_{\G_{t_k}}\Hb\parM_{hr}\right), \\ 
\tilde{\Hb} &:=\tanh \left(\Lb_{\G_{t_k}}\Xb_{t_k}\parM_{xh} + \Lb_{\G_{t_k}}\left(\mathbf{R} \odot \Hb\right)\parM_{hh}\right). 
\end{aligned}
\end{equation}
Finally, the \textit{post--jump} node features are obtained as
\begin{equation}
\begin{aligned} 
\Hb^+ &={\tt GCGRU}(\Hb,\Xb_t):=\mathbf{Z} \odot \Hb+(\mathbb{1}-\mathbf{Z}), \odot \tilde{\Hb} \end{aligned}
\end{equation}
where $\parM_{xz},~\parM_{hz},~\parM_{xr},~\parM_{hr},~\parM_{xh},~\parM_{hh}$ are matrices of trainable parameters, $\sigma$ is the standard sigmoid activation and $\mathbb{1}$ is all--ones matrix of suitable dimensions.

\section{Additional experimental details}

\paragraph{Computational resources}
We carried out all experiments on a cluster of 4x12GB NVIDIA\textsuperscript{\textregistered} Titan Xp GPUs and CUDA 10.1. The models were trained on GPU.

\subsection{Node Classification}
\paragraph{Training hyperparameters}
All models are trained for $2000$ epochs using Adam~\citep{kingma2014adam} with learning rate $lr = 10^{-3}$ on Cora, Citeseer and $lr = 10^{-2}$ on Pubmed due to its training set size. The reported results are obtained by selecting the lowest validation loss model after convergence (i.e. in the epoch range $1000$ -- $2000$). Test metrics are not utilized in any way during the experimental setup. For the experiments to test resilience to integration time changes, we set a higher learning rate for all models i.e. $lr = 10^{-2}$ to reduce the number of epochs necessary to converge.

\paragraph{Architectural details}
\textit{SoftPlus} is used as activation for GDEs. Smooth activations have been observed to reduce stiffness \citep{chen2018neural} of the ODE and therefore the number of function evaluations (NFE) required for a solution that is within acceptable tolerances. All the other activation functions are \textit{rectified linear units} (ReLU). The exact input and output dimensions for the GCDE architectures are reported in Table \ref{tab:gde_arch}. The vector field $\Fb$ of GCDEs--rk2 and GCDEs--rk4 is parameterized by two GCN layers. GCDEs--dopri5 shares the same structure without GDE--2 (GCN). Input GCN layers are set to dropout $0.6$ whereas GCN layers parametrizing $\Fb$ are set to $0.9$.
\begin{table}%{\columnwidth}
\centering
\setlength\tabcolsep{3pt}
\begin{tabular}[t]{lrrr}  
\toprule
Layer & Input dim. & Output dim. & Activation \\
\midrule
GCN--in & dim. in & $64$ & ReLU \\
GDE--1 (GCN) & $64$ & $64$ & Softplus  \\
GDE--2 (GCN) & $64$ & $64$ & None \\
GCN-out & 64 & dim. out & None \\
\bottomrule

\end{tabular}
\vspace{2mm}
\caption{General architecture for GCDEs on node classification tasks. GCDEs applied to different datasets share the same architecture. The vector field $\Fb$ is parameterized by two GCN layers. GCDEs--dopri5 shares the same structure without GDE--2 (GCN).}
\label{tab:gde_arch}
\end{table}

\begin{figure*}[t]
    \centering
    \includegraphics[width=1\linewidth]{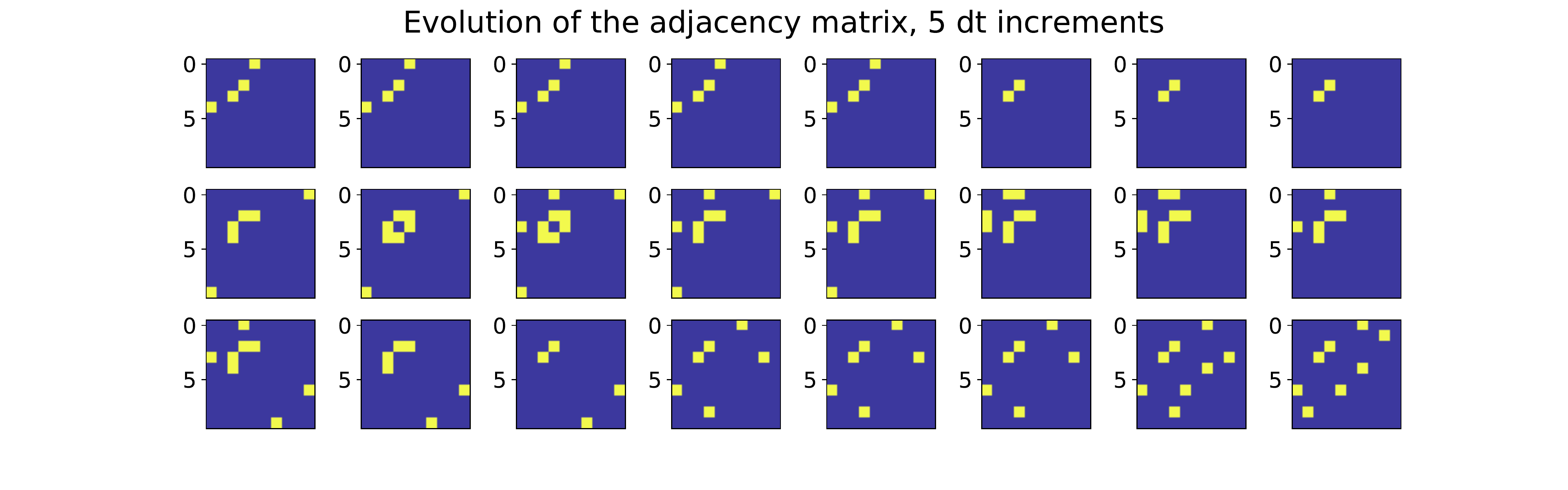}
    \vspace{-5mm}
    \caption{Snapshots of the evolution of adjacency matrix $\Ab_t$ throughout the dynamics of the multi--particle system. Yellow indicates the presence of an edge and therefore a reciprocal force acting on the two bodies}
    \label{fig:adj}
\end{figure*}

\subsection{Multi--Agent System Dynamics}
\paragraph{Dataset}
Let us consider a planar multi agent system with states $\x_i$ ($i = 1,\dots,n$) and second--order dynamics:
\begin{align*}
    \ddot \x_i &= -\x_i - \sum_{j\in\N_i}\f_{ij}(\x_i,\x_j,\dot\x_i,\dot\x_j),
\end{align*}
where 
\begin{align*}
    &\f_{ij} = -\left[\alpha\lc \|\x_i - \x_j\| - r \rc + \beta\frac{\langle\dot\x_i - \dot\x_j, \x_i - \x_j\rangle}{\|\x_i - \x_j\|}\right]\mathbf{n}_{ij},\\
    &\mathbf{n}_{ij} = \frac{\x_i - \x_j}{\|\x_i - \x_j\|},\quad \alpha,~\beta,~r > 0,
\end{align*}
and
\[
    \quad \N_i := \left\{j:2\|\x_i-\x_j\|\leq r \land j\not = i\right\}.
\]
The force $\f_{ij}$ resembles the one of a spatial spring with drag interconnecting the two agents. The term $-\x_i$, is used instead to stabilize the trajectory and avoid the "explosion" of the phase--space. Note that $\f_{ij} = -\f_{ji}$. The adjaciency matrix $\Ab_t$ is computed along a trajectory
\[
    \Ab_t^{(ij)} = 
    \left\{\begin{matrix*}[l]
        1 & 2\|\x_i(t)-\x_j(t)\|\leq r\\
        0 & \text{otherwise}
    \end{matrix*}\right.,
\]
which indeed results to be symmetric, $\Ab_t = \Ab_t^\top$ and thus yields an undirected graph. Figure \ref{fig:adj} visualizes an example trajectory of $\Ab_t$.
\begin{table*}%{\columnwidth}
\centering
\setlength\tabcolsep{3pt}
\begin{tabular}[t]{lrrrrrrr}  
\toprule
Model & MAPE$_{1}$ & MAPE$_{3}$ & MAPE$_{5}$ & MAPE$_{10}$ & MAPE$_{15}$ & MAPE$_{20}$ & MAPE$_{50}$ \\
\midrule
Static & $26.12$ & $160.56$ & $197.20$ & $ 235.21$ & $261.56$ & $275.60$ & $360.39$ \\
Neural ODE  & $ 26.12$ & $52.26$ & $92.31$ & $ 156.26$ & $238.14$ & $301.85$ & $668.47$  \\
GDE & $13.53$ & $15.22$ & $18.76$ & $27.76$ & $33.90$ & $42.22$ & $77.64$ \\
GDE--II & $13.46$ & $14.75$ & $17.81$ & $27.77$ & $32.28$ & $40.64$ & $73.75$  \\
\bottomrule
\end{tabular}
\caption{Mean MAPE results across the 10 multi--particle dynamical system experiments. MAPE$_i$ indicates results for $i$ extrapolation steps on the full test trajectory.}
\label{tab:mape_table}
\end{table*}
\begin{wrapfigure}[25]{r}{0.45\textwidth}
    \centering
    \includegraphics[scale=0.4]{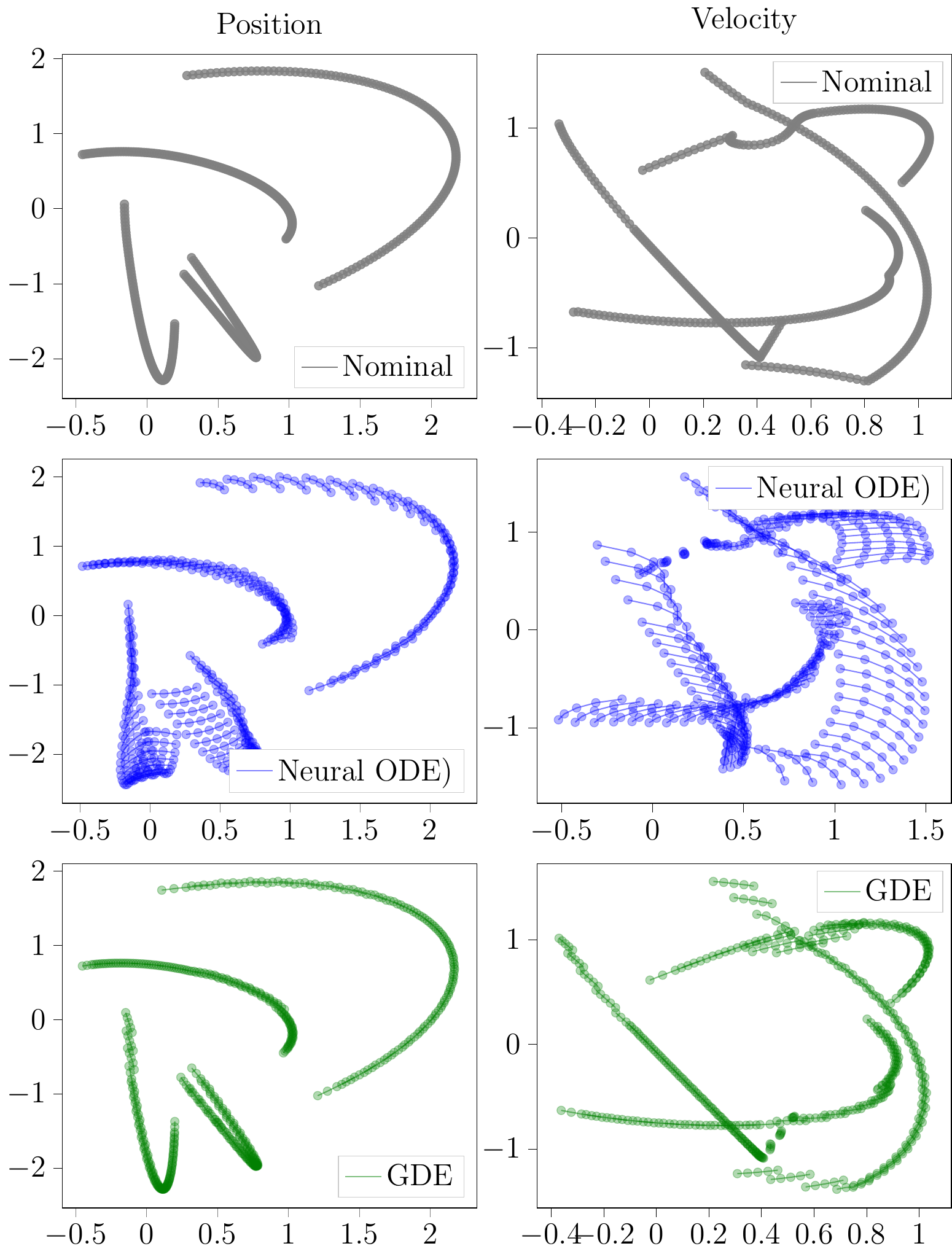}
    \caption{Test extrapolation, $5$ steps. Trajectory predictions of Neural ODEs and GDEs. The extrapolation is terminated after 5 steps and the nominal state is fed to the model.}
    \label{fig:extrapo_traj}
\end{wrapfigure}
We collect a single rollout with $T = 5$, $dt = 1.95\cdot10^{-3}$ and $n = 10$. The particle radius is set to $r = 1$.
\paragraph{Architectural details}
Node feature vectors are $4$ dimensional, corresponding to the dimension of the state, i.e. position and velocity. Neural ODEs and \textit{Static} share an architecture made up of 3 fully--connected layers: $4n$, $8n$, $8n$, $4n$ where $n = 10$ is the number of nodes. The last layer is linear. We evaluated different hidden layer dimensions: $8n$, $16n$, $32n$ and found $8n$ to be the most effective. Similarly, the architecture of first order GCDEs is composed of 3 GCN layers: $4$, $16$, $16$, $4$. Second--order GCDEs, on the other hand, are augmented by $4$ dimensions: $8$, $32$, $32$, $8$. We experimented with different ways of encoding the adjacency matrix $\Ab$ information into Neural ODEs and $Static$ but found that in all cases it lead to worse performance. 
\paragraph{Additional results}
We report in Figure~\ref{fig:extrapo_traj} test extrapolation predictions of $5$ steps for GDEs and the various baselines. Neural ODEs fail to track the system, particularly in regions of the state space where interaction forces strongly affect the dynamics. GDEs, on the other hand, closely track both positions and velocities of the particles.

% to their inability to effectively incorporate structural inductive biases. 
% %  Neural ODEs

%  for long extrapolations of the trajectory, whereas Neural ODEs diverge. GCDE-II

%

% \begin{figure}[16]{r}{0.45\textwidth}
%     \centering
%     \includegraphics[width=1\linewidth]{figures/ma_extrapo2.pdf}
%     \caption{Test extrapolation, 5 steps. Trajectory predictions of Neural ODEs and GDEs. The extrapolation is terminated after 5 steps and the nominal state is fed to the model, as described in the experimental setup.}
%     \label{fig:extrapo_traj}
% \end{figure}
% \end{wrapfigure}

\subsection{Traffic Forecasting}
\paragraph{Dataset and metrics}
The timestamp differences between consecutive graphs in the sequence varies due to undersampling. The distribution of timestamp deltas (5 minute units) for the three different experiment setups (30\%, 50\%, 70\% undersampling) is shown in Figure \ref{fig:traffic_delta_ts}.

As a result, GRU takes 230 dimensional vector inputs (228 sensor observations + 2 additional features) at each sequence step. Both GCGRU and GCDE--GRU graph inputs with and 3 dimensional node features (observation + 2 additional feature). The additional time features are excluded for the loss computations.
We include MAPE and RMSE test measurements, defined as follows:
\begin{figure*}[t]
    \centering
    \includegraphics{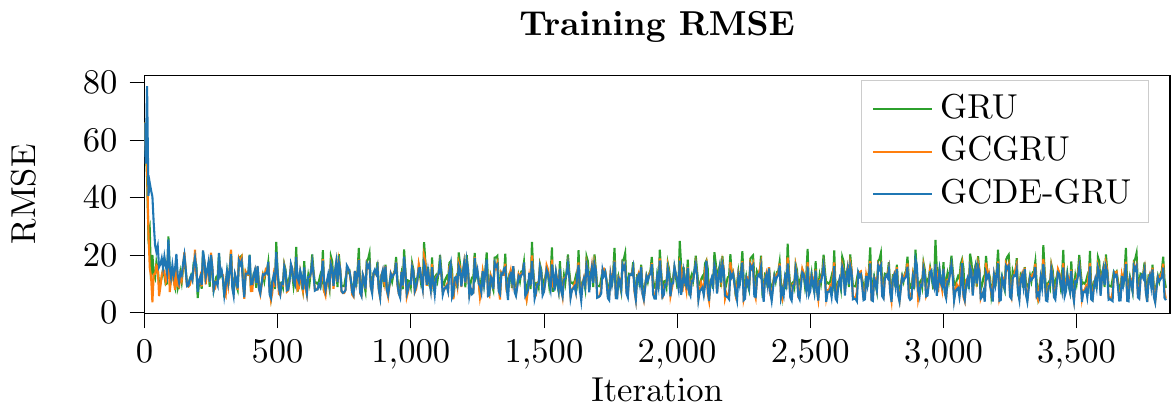}
    \caption{Traffic data training results of 50\% undersampling.}
    \label{fig:traffic_train}
\end{figure*}
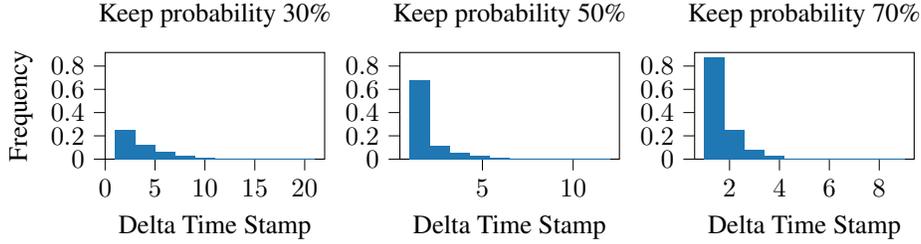
\begin{figure*}[h!]
    \centering
    \begin{tikzpicture}

\definecolor{color0}{rgb}{0.12156862745098,0.466666666666667,0.705882352941177}

\begin{groupplot}[group style={group size=3 by 1}]
\nextgroupplot[
width = 4.5 cm,
height = 3cm,
tick align=outside,
tick pos=left,
title={Keep probability 30\%},
x grid style={white!69.0196078431373!black},
xlabel={Delta Time Stamp},
xmin=0, xmax=22,
xtick style={color=black},
y grid style={white!69.0196078431373!black},
ylabel={Frequency},
ymin=0, ymax=0.916538592896175,
ytick style={color=black}
]
\draw[draw=none,fill=color0] (axis cs:1,0) rectangle (axis cs:3,0.253164556962025);
\draw[draw=none,fill=color0] (axis cs:3,0) rectangle (axis cs:5,0.124472573839662);
\draw[draw=none,fill=color0] (axis cs:5,0) rectangle (axis cs:7,0.0638185654008439);
\draw[draw=none,fill=color0] (axis cs:7,0) rectangle (axis cs:9,0.0313818565400844);
\draw[draw=none,fill=color0] (axis cs:9,0) rectangle (axis cs:11,0.01292194092827);
\draw[draw=none,fill=color0] (axis cs:11,0) rectangle (axis cs:13,0.00659282700421941);
\draw[draw=none,fill=color0] (axis cs:13,0) rectangle (axis cs:15,0.00395569620253165);
\draw[draw=none,fill=color0] (axis cs:15,0) rectangle (axis cs:17,0.00210970464135021);
\draw[draw=none,fill=color0] (axis cs:17,0) rectangle (axis cs:19,0.000791139240506329);
\draw[draw=none,fill=color0] (axis cs:19,0) rectangle (axis cs:21,0.000791139240506329);

\nextgroupplot[
width = 4.5cm,
height = 3cm,
tick align=outside,
tick pos=left,
title={Keep probability 50\%},
x grid style={white!69.0196078431373!black},
xlabel={Delta Time Stamp},
xmin=0.45, xmax=12.55,
xtick style={color=black},
y grid style={white!69.0196078431373!black},
ymin=0, ymax=0.916538592896175,
ytick style={color=black}
]
\draw[draw=none,fill=color0] (axis cs:1,0) rectangle (axis cs:2.1,0.682529743268629);
\draw[draw=none,fill=color0] (axis cs:2.1,0) rectangle (axis cs:3.2,0.117265327033643);
\draw[draw=none,fill=color0] (axis cs:3.2,0) rectangle (axis cs:4.3,0.0537940456537826);
\draw[draw=none,fill=color0] (axis cs:4.3,0) rectangle (axis cs:5.4,0.0281778334376957);
\draw[draw=none,fill=color0] (axis cs:5.4,0) rectangle (axis cs:6.5,0.0128081061080435);
\draw[draw=none,fill=color0] (axis cs:6.5,0) rectangle (axis cs:7.6,0.00626174076393237);
\draw[draw=none,fill=color0] (axis cs:7.6,0) rectangle (axis cs:8.7,0.00313087038196618);
\draw[draw=none,fill=color0] (axis cs:8.7,0) rectangle (axis cs:9.8,0.00313087038196619);
\draw[draw=none,fill=color0] (axis cs:9.8,0) rectangle (axis cs:10.9,0.00113849832071498);
\draw[draw=none,fill=color0] (axis cs:10.9,0) rectangle (axis cs:12,0.000853873740536233);

\nextgroupplot[
width = 4.5cm,
height = 3cm,
tick align=outside,
tick pos=left,
title={Keep probability 70\%},
x grid style={white!69.0196078431373!black},
xlabel={Delta Time Stamp},
xmin=0.6, xmax=9.4,
xtick style={color=black},
y grid style={white!69.0196078431373!black},
ymin=0, ymax=0.916538592896175,
ytick style={color=black}
]
\draw[draw=none,fill=color0] (axis cs:1,0) rectangle (axis cs:1.8,0.872893897996357);
\draw[draw=none,fill=color0] (axis cs:1.8,0) rectangle (axis cs:2.6,0.255009107468124);
\draw[draw=none,fill=color0] (axis cs:2.6,0) rectangle (axis cs:3.4,0.0836748633879781);
\draw[draw=none,fill=color0] (axis cs:3.4,0) rectangle (axis cs:4.2,0.0281762295081967);
\draw[draw=none,fill=color0] (axis cs:4.2,0) rectangle (axis cs:5,0);
\draw[draw=none,fill=color0] (axis cs:5,0) rectangle (axis cs:5.8,0.00711520947176684);
\draw[draw=none,fill=color0] (axis cs:5.8,0) rectangle (axis cs:6.6,0.00199225865209472);
\draw[draw=none,fill=color0] (axis cs:6.6,0) rectangle (axis cs:7.4,0.000569216757741348);
\draw[draw=none,fill=color0] (axis cs:7.4,0) rectangle (axis cs:8.2,0.000284608378870674);
\draw[draw=none,fill=color0] (axis cs:8.2,0) rectangle (axis cs:9,0.000284608378870674);
\end{groupplot}

\end{tikzpicture}
    \caption{Distribution of deltas between timestamps $t_{k+1} - t_k$ in the undersampled dataset. The time scale of required predictions varies greatly during the task.}
    \label{fig:traffic_delta_ts}
\end{figure*}
\begin{equation}
    \text{MAPE}(\yb, \hat{\yb}) = \frac{100 \%}{pT}\norm{\sum_{t=1}^{T}(\y_t - \hat{\y}_t) \oslash \y_t}_1,
\end{equation}
where $\yb, \text{and} \ \hat{\yb} \in \mathbb R^p $ is the set of vectorized target and prediction of models respectively. $\oslash$ and $\norm{\cdot}_1$ denotes Hadamard division and the 1-norm of vector. 

\begin{align*}
    \text{RMSE}(\yb, \hat{\yb}) &= \frac{1}{p}\norm{\sqrt{\frac{1}{T}\sum_{t=1}^{T}  (\y_t - \hat{\y}_t)^2}}_1,
\end{align*}

where $(\cdot)^2$ and $\sqrt{\cdot}$ denotes the element-wise square and square root of the input vector, respectively. $\yb_t \ \text{and} \ \hat{\yb}_t$ denote the target and prediction vector. 
\begin{figure*}[t]
    \centering
    \input{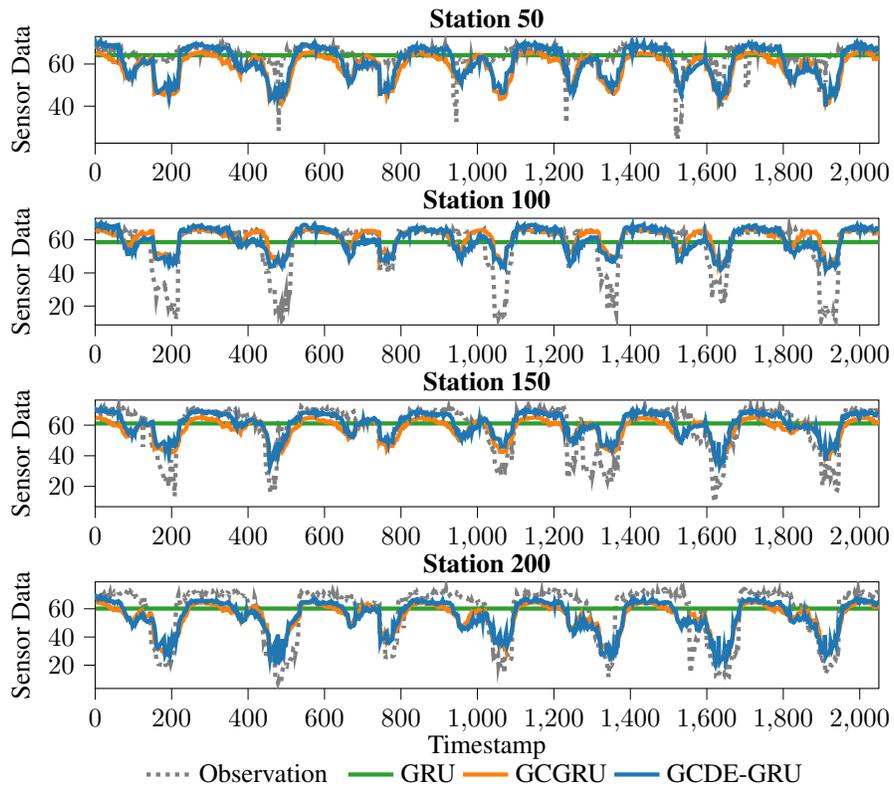}
    \caption{Traffic data prediction results of 50\% undersampling. GCDE--GRUs are able to evolve the latents between timestamps and provide a more accurate fit.}
    \label{fig:traffic_preds}
\end{figure*}
\paragraph{Architectural details}
We employed two baseline models for contextualizing the importance of key components of GCDE--GRU. \textit{GRUs} architectures are equipped with 1 GRU layer with hidden dimension 50 and a 2 layer fully--connected head to map latents to predictions. \textit{GCGRUs} employ a GCGRU layer with 46 hidden dimension and a 2 layer fully--connected head. Lastly, GCDE--GRU shares the same architecture GCGRU with the addition of the flow $\Fb$ tasked with evolving the hidden features between arrival times. $\Fb$ is parametrized by 2 GCN layers, one with tanh activation and the second without activation. ReLU is used as the general activation function.

\paragraph{Training hyperparameters}
All models are trained for 40 epochs using Adam\citep{kingma2014adam} with $lr = 10^{-2}$. We schedule $lr$ by using cosine annealing method \citep{loshchilov2016sgdr} with $T_0=10$. The optimization is carried out by minimizing the \textit{mean square error} (MSE) loss between predictions and corresponding targets. 

\paragraph{Additional results}
Training curves of the models are presented in the Fig \ref{fig:traffic_train}. All of models achieved nearly 13 in RMSE during training and fit the dataset. However, due to the lack of dedicated spatial modeling modules, GRUs were unable to generalize to the test set and resulted in a mean value prediction.

%
% \begin{table*}[t]
% \small
% \centering
% \setlength\tabcolsep{3pt}
% \begin{tabular}{lrrr}
% \toprule
% Model (depth) & Cora & Citeseer & Pubmed\\ %& NFE \\
% \midrule
% GCN(2) & $81.4 \pm 0.5\%$ & $70.9 \pm 0.5\%$ & $79.0 \pm 0.3\%$ & 184 -- 474 -- 64 \\%& --  \\

% GCN^*(2) & $82.8 \pm 0.3\%$ & $71.2 \pm 0.4\%$ & $79.5 \pm 0.4\%$ & 184 -- 474 -- 64 \\%& --  \\

% % rGCN(3) & $81.4 \pm 0.5\%$ & $70.9 \pm 0.5\%$ & $79.0 \pm 0.3\%$ & 184 -- 474 -- 64 & --  \\
% %GAT (2) & $83.0 \pm 0.7\%$ & $\textbf{72.5} \pm 0.7\%$ & $79.0 \pm 0.3\%$ & 93k/238k/39k  \\
% \midrule
% GCDE--rk2 (2) & $83.0 \pm 0.6\%$ &  $72.3 \pm 0.5\%$ & $\textbf{79.9} \pm 0.3\%$ & 188 -- 478 -- 68 \\%& $2$ -- $2$ -- $2$ \\
% GCDE--rk4 (4) & $\textbf{83.8} \pm 0.5\%$ &  $\textbf{72.5} \pm 0.5\%$ & $79.5 \pm 0.4\%$ & 188 -- 478 -- 68 \\%& $4$ -- $4$ -- $4$\\
% GCDE--dpr5 \textbf{(158)} & $81.8 \pm 1.2\%$ & $68.3\pm 1.2\%$ & $78.5 \pm 0.7\%$ & 188 -- 478 -- 68 \\%& $236$ -- $257$ -- $167$ \\
% %CGCDE--dpr5 \textbf{(120)} & $82.0 \pm 1.1\%$ & $70.5\pm 1.0\%$ & $78.6 \pm 0.9\%$ & 192 -- 482 -- 72 \\%& $223$ -- $240$ -- $150$  \\
% \bottomrule
% \end{tabular}
% \caption{Test results in percentages across 100 runs (mean and standard deviation). All models have hidden dimension set to $64$. For the baseline GCN result we refer to \citep{velivckovic2017graph}. We report model size as the number of parameters in thousands (k) for each of the three datasets.}
% \label{tab:allresone}
% \end{table*}

\end{document}